\documentclass{article}
\usepackage{ijcai25}

\usepackage{times}
\usepackage{soul}
\usepackage{url}
\usepackage[hidelinks]{hyperref}
\usepackage[utf8]{inputenc}
\usepackage[small]{caption}
\usepackage{graphicx}
\usepackage{amsmath}
\usepackage[dvipsnames]{xcolor}
\usepackage{amsthm}
\usepackage{booktabs}
\usepackage{algorithm}
\usepackage{algorithmic}
\usepackage[switch]{lineno}
\usepackage{tikz}
\usetikzlibrary{shapes.geometric, arrows.meta, positioning}

\usepackage[capitalize]{cleveref}

\usepackage{amssymb}
\usepackage{xcolor}
\usepackage{empheq}
\usepackage{natbib}

\title{Counterfactual Routing Using Integer Programming with Constraint Generation}

\author{
Dani\"el Vos
\and
Sterre Lutz
\affiliations
Delft University of Technology\\
\emails
\{S.Lutz, D.A.Vos\}@tudelft.nl
}

\begin{document}

\maketitle

\begin{abstract}
We present our submission to the IJCAI 2025 `Counterfactual
Routing Competition' (CRC 25). The goal of the competition is to find counterfactual explanations for the shortest path problem. 
This requires deciding what the minimal changes to a road network would make a route chosen by the user the optimal route. This enables explanations such as ``Your suggested route \textit{would} indeed have been optimal, if road X were not a bicycle path.''
Our solution models the problem as an integer program, iteratively incorporating constraints until an exact solution is found. 
In the final evaluation on held-out test instances, our method ranked fourth in solution quality and obtained its solution fastest on every instance, with an average runtime of 9.0 seconds compared to 118.8 seconds for the next-fastest submission.
\end{abstract}
\paragraph{Code} \url{https://github.com/SUMI-lab/CRC-competition}
\section{Introduction}
We present our solution to the IJCAI 2025 `Counterfactual Routing Competition' (CRC 25). 
The competition addresses the challenge of safe and accessible navigation for wheelchair users, who face unique obstacles beyond simply reaching their destination quickly. 
Factors such as curb heights, sidewalk widths, and crossings significantly impact which routes are suitable. 

While existing routing algorithms can generate optimal paths considering these constraints, they often do not explain why a particular route is chosen over others. 
The competition organizers emphasize that this lack of transparency can undermine users’ trust in the recommended routes.

To address this, the competition requires participants to create counterfactual explanations that demonstrate why a chosen ‘foil’ route is suboptimal. Such explanations show how an alternative route would have been optimal if some road conditions had been different. Providing counterfactuals for routing requires identifying minimal changes in complex, weighted graphs representing the road maps.

Our solution models the problem as an integer program that iteratively incorporates constraints to ensure the counterfactual route becomes optimal after minimal modifications to the road map in question.

\paragraph{Competition results} In the final competition evaluation\footnote{\url{https://sites.google.com/view/crc25-ijcai/competition-results}} on the held-out test instances, our submission ranked fourth in terms of solution quality. Notably, our method was the fastest on every single test instance, with an average runtime of 9.0 seconds compared to 118.8 seconds for the next-fastest submission.

\paragraph{Related Work}

We did not directly base our solution on other work; however, we applied the general idea of cutting plane methods for integer programs to the competition (\citet{bradley1977applied}, Chapter 9.8). This approach is useful, for example, when an (integer) linear program is too large to solve directly. Instead of solving the whole program, cuts are added iteratively until optimality has been proven.

Our use of integer programming was partly motivated by \citet{DBLP:conf/icml/KantchelianTJ16}. They use an integer program to find optimal adversarial examples, which is nearly equivalent to finding optimal counterfactual explanations, i.e., they minimize distance while ensuring a specific outcome.
\section{Problem Description}
We are given a segment of the road map of the city of Amsterdam, and a user-defined `foil' route. The problem to solve is: make minimal changes to the map such that the user's route becomes the shortest one from point A to point B.

Formally, the map of the neighborhood is defined as the multi-graph $\mathcal{G} = \{\mathcal{V}, \mathcal{E}\}$, where $\mathcal{V}$ denotes the set of nodes (i.e., locations), and $\mathcal{E}$ denotes the set of edges (i.e., roads or paths connecting these locations). Each edge contains some attributes such as its length, width, curb height and path type. The user's foil route is a path of edges $r^t$, and the user's preferences are captured by user model $U$.
The high-level optimization problem given a route planner $\pi_U^r$ under the user model $U$ is:
\begin{align*}
\min_{\mathcal{G}^c} \quad & d_g(\mathcal{G}, \mathcal{G}^c)
\\
\text{s.t.} \quad & d_r\left(r^t, \pi_U^r(\mathcal{G}^c)\right) \leq \delta,
\end{align*}
here $\mathcal{G}^c$ is the desired counterfactual map (output), $d_g$ represents the graph difference, $\pi_U^r(\mathcal{G}^c)$ is the route returned by the planner for the given user model and the counterfactual map, $d_r$ denotes the route difference, and $\delta$ is a predefined threshold fixed to 0.05 in this competition.

\subsection{Concrete Optimization Problem} \label{sec:optimization-problem}

The choice of graph distance function $d_g$ and route distance function $d_r$ greatly affects the tractability of the optimization problem. Therefore, it is important to specify the problem with concrete distances and objectives. 

In this competition, the planner $\pi_U^r$ simply finds the minimum weight path between point A and B using the Dijkstra algorithm. The weight of each edge $e \in \mathcal{E}$ is determined as:
\begin{equation} \label{eq:edge-weight}
    w_{e} = \text{length}_e \times U_\text{crossing}^{\mathbb{I}[e \text{ is crossing}]} \times U_\text{preference bonus}^{\mathbb{I}[e_\text{type} = U_\text{preference}]},
\end{equation}
with typical values for the weights being $U_\text{crossing} = 1.4$ and $U_\text{preference bonus} = 0.6$.
There is another important requirement for edges, that is, a user will only take them if both: their minimum width is greater than or equal to $U_\text{min width}$ \textbf{and} their maximum curb height is less than or equal to $U_\text{max curb}$.
This means that, effectively, the weight of an edge can be thought of as infinite if these predicates do not hold.

The graph distance $d_g = |\mathcal{OP}|_\mathcal{G}^{\mathcal{G}'}$ is measured by the amount of operations ($|\mathcal{OP}|$) that are needed to transform $\mathcal{G}$ into $\mathcal{G}'$. Here, the allowed operations for each edge $e$ are:
\begin{itemize}
    \item Change the type of $e$
    \item Change the minimum width of $e$
    \item Change the maximum curb height of $e$
\end{itemize}
These operations modify the weights of the edges, which in turn can impact the minimum-weight path in the graph.

Finally, the route distance function is given by the weighted common edge rate
\begin{equation*}
    d_r(r, r') = 1 - 2 \frac{\sum_{e \in r \cap r'} \text{length}_e}{\sum_{e \in r} \text{length}_e + \sum_{e \in r'} \text{length}_e},
\end{equation*}
which intuitively measures the lengths of the edges on which the paths overlap.

While the competition allows a slack of 0.05 in the route distance $d_r$ between the shortest path in the counterfactual graph $\mathcal{G}'$ and the user's foil route, we disallow any slack in our approach. This choice allows for a simpler tractable formulation and can be made equivalent to the optimization problem with slack by enumerating all paths $r$ such that $d_r(r^t, r') \leq 0.05$ (see Section~\ref{sec:future-work}). The concrete optimization problem that we solve is:
\begin{align*}
\min_{\mathcal{G}^c} \quad & |\mathcal{OP}|_\mathcal{G}^{\mathcal{G}^c} & {\color{gray} \triangleright\text{ minimize number of changes}}
\\
\text{s.t.} \quad & \pi_U^r(\mathcal{G}^c) = r_t. & {\color{gray} \triangleright\text{ s.t. the user route is shortest}}
\end{align*}
\section{Method Description} \label{sec:method}

At a high level, we will solve the problem using an integer program to which we iteratively add constraints during the solve via callbacks. For our implementation, we chose to use GUROBI\footnote{\url{https://www.gurobi.com/}} 12.0.2, but any integer programming solver that allows for constraint generation can be used.

Let us first introduce the integer program without constraint generation. Our formulation solves the problem at a slightly abstract, more intuitive level by reasoning over the effects of the operations $\mathcal{OP}$ that we are allowed to perform. Specifically, the operations allow us to make the following changes to an edge $e$ of the input graph $\mathcal{G}$:
\begin{itemize}
    \item \textbf{Decrease or increase the edge weight by a factor $U_\text{preference bonus}$}, depending on the edge type.
    \item \textbf{Disable an enabled edge} by decreasing the width of a segment or increasing the maximum curb height. Sometimes this operation is not possible.
    \item \textbf{Enable a disabled edge} by increasing the width of a segment or reducing the maximum curb height. Sometimes both operations are necessary.
\end{itemize}
To this end, we introduce three binary variables for each edge $e \in \mathcal{E}$, $\mathbf{z}^\text{cost}_e, \mathbf{z}^\text{disable}_e \text{ and } \mathbf{z}^\text{enable}_e$, indicating whether to change the weight, enable the edge, or disable the edge. For clarity, we will use bold symbols to denote decision variables and regular symbols for constants. Whenever it is not possible to enable or disable an edge $e$, we constrain $\mathbf{z}^\text{enable}=0$ or $\mathbf{z}^\text{disable}=0$.

The objective is to minimize the sum of the $\mathbf{z}$ variables since these correspond to the number of operations $|\mathcal{OP}|$ that we make to the input graph. It is important to note, however, that the mapping between $\sum \mathbf{z}$ and $|\mathcal{OP}|$ is not exactly one-to-one as it might be necessary to change both the height of a curb and the width of a road segment to enable the user to take it (see Section~\ref{sec:optimization-problem}). Therefore, we introduce constants $C_e$ that hold the number of operations needed to enable an edge. Disabling an edge or changing its weight can always be done in one operation. The resulting objective function is:
\begin{equation*}
    \min_{\mathbf{z}} \ \sum_e \mathbf{z}^\text{cost}_e + \mathbf{z}^\text{disable}_e + C_e \mathbf{z}^\text{enable}_e.
\end{equation*}

Next, we must constrain the model to only allow values for $\mathbf{z}$ that together ensure the user's foil route is the path with the least weight in the graph. We introduce the notation $\mathcal{P}_\mathcal{G}(s, t)$ to denote the set of all simple paths in the graph $\mathcal{G}$ starting in node $s$ and ending in node $t$. Now, to constrain the foil path $P_\text{foil}$ to be the one with least weight is equivalent to forcing it to be weighted (strictly) less than any other simple path:
\begin{equation*}
    W(P_\text{foil}, \mathbf{z}) \leq W(P, \mathbf{z}) - \epsilon \quad \forall P \in \mathcal{P}_\mathcal{G}(s, t) {\setminus} \{P_\text{foil}\},
\end{equation*}
where $W(P', \mathbf{z})$ is the weight of a path $P'$ given the graph with changes induced by $\mathbf{z}$, and $\epsilon$ is a small constant to enforce strict inequality. 
Notice that $W(P', \mathbf{z})$ can be written as a linear expression:
\begin{equation*}
    W(P, \mathbf{z}) = \sum_{e \in P} w_e + \Delta_e z_e^\text{cost} + M z_e^\text{disable} + M(1 - z_e^\text{enable}),
\end{equation*}
where $w_e$ refers to the weight of edge $e$ without changes to the graph (Equation~\ref{eq:edge-weight}), $\Delta_e$ is the weight change incurred by changing the type of $e$, and $M$ is a large constant. It is important to choose $M$ large enough to trivially satisfy the constraints when the appropriate edges are disabled/enabled, but as small as possible to prevent numerical issues in the solver. We select $M = \epsilon + \frac{1}{U_\text{preference bonus}} \sum_{e \in P_\text{foil}} w_e$, which ensures that any path with an edge of weight $M$ is larger than the weight of foil path $P_\text{foil}$ even if all its edges were changed into the user's unpreferred type.

Since both our constraints and objective are linear, the combined program is an exponentially-sized integer linear program (ILP) that solves the problem:
\begin{empheq}[box=\fbox]{align*}
    \min_{\mathbf{z}} \;\; & \sum_e \mathbf{z}^\text{cost}_e + \mathbf{z}^\text{disable}_e + C_e \mathbf{z}^\text{enable}_e  \\
    \text{s.t.} \;\; & W(P_\text{foil}, \mathbf{z}) \leq W(P, \mathbf{z}) - \epsilon \;\; \forall P {\in} \mathcal{P}_\mathcal{G}(s, t) {\setminus} \{P_\text{foil}\} \\
    & \mathbf{z}^\text{cost}_e, \mathbf{z}^\text{disable}_e, \mathbf{z}^\text{enable}_e \in \{0, 1\} \;\; \forall e \in \mathcal{E}.
\end{empheq}

\subsection{Constraint Generation}

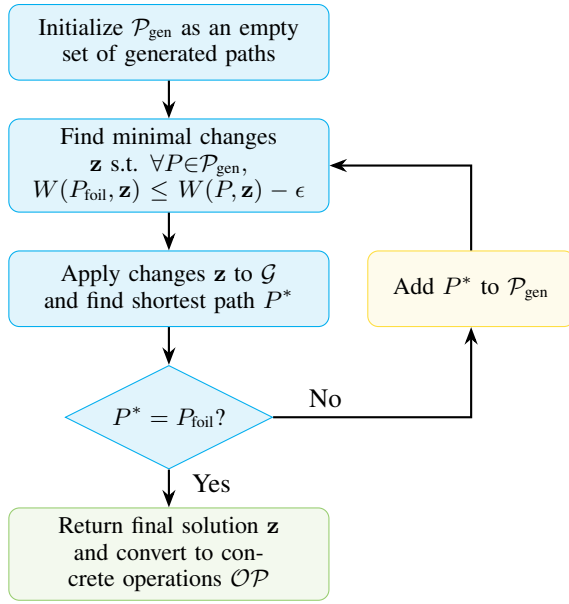
\begin{figure}
\centering
\tikzstyle{process} = [rectangle, rounded corners, minimum height=1cm, text centered, font=\small, text width=4cm, align=center]
\tikzstyle{decision} = [diamond, aspect=2, minimum height=1cm, text centered, draw=black, fill=white, font=\small, text width=1.5cm, align=center]
\tikzstyle{arrow} = [thick,->,>=Stealth]

\begin{tikzpicture}[node distance=0.5cm]

\node (start) [process, draw=cyan, fill=cyan!10] 
{Initialize $\mathcal{P}_{\text{gen}}$ as an empty set of generated paths};
\node (solve) [process, below=of start, draw=cyan, fill=cyan!10] 
{Find minimal changes $\mathbf{z}$ s.t. $\forall P {\in} \mathcal{P}_{\text{gen}}, $\\$W(P_\text{foil}, \mathbf{z}) \leq W(P, \mathbf{z}) - \epsilon$};
\node (extract) [process, below=of solve, draw=cyan, fill=cyan!10] 
{Apply changes $\mathbf{z}$ to $\mathcal{G}$ and find shortest path $P^*$};
\node (check) [decision, below=of extract, draw=cyan, fill=cyan!10] {$P^* = P_\text{foil}$?};
\node (add) [process, right=of extract, draw=Goldenrod, fill=Goldenrod!10, text width=2.5cm] 
{Add $P^*$ to $\mathcal{P}_{\text{gen}}$};
\node (end) [process, below=of check, draw=YellowGreen, fill=YellowGreen!10] 
{Return final solution $\mathbf{z}$ and convert to concrete operations $\mathcal{OP}$};

\draw [arrow] (start) -- (solve);
\draw [arrow] (solve) -- (extract);
\draw [arrow] (extract) -- (check);
\draw [arrow] (check) -| node[above left,xshift=-45pt] {No} (add);
\draw [arrow] (add) |- (solve);
\draw [arrow] (check) -- node[right,xshift=5pt,yshift=2pt] {Yes} (end);

\end{tikzpicture}
\caption{Flowchart showing the iterative process of the proposed constraint generation method. We start with an empty formulation and add shortest path constraints until the user's foil route becomes the shortest one. }
\label{fig:diagram}
\end{figure}

While the aforementioned ILP solves the problem, its formulation is potentially exponential in size due to the enumeration of all simple paths between the source and target nodes $\mathcal{P}_\mathcal{G}(s, t)$. To arrive at a tractable formulation, we only add the necessary constraints via a constraint generation process. A schematic of this process is given in \cref{fig:diagram}.

Such a constraint generation process is implemented via solver callbacks that are triggered when the solver finds an integer assignment to the variables $\mathbf{z}$ that satisfy the current set of constraints. In each callback we then either add a constraint that invalidates the proposed solution and continue solving, or accept the proposed solution. To identify the violating constraint we simply apply the operations to input graph $\mathcal{G}$ as indicated by values of $\mathbf{z}$ and solve a shortest path problem using the polynomial-time Dijkstra algorithm.
Let $\mathcal{G}^{\mathbf{z}}$ denote the input graph with changes applied as indicated by the values of $\mathbf{z}$, then we identify the least-weight path:
\begin{equation*}
    P^* = \arg\min_{P \in \mathcal{P}_{\mathcal{G}^{\mathbf{z}}}(s, t)} \sum_{e \in P} w_e.
\end{equation*}
If $P^* = P_\text{foil}$ then we know $P_\text{foil}$ is the shortest route in the graph and therefore the problem is solved. Instead, if $P^* \neq P_\text{foil}$, we add the constraint
\begin{equation*}
    W(P_\text{foil}, \mathbf{z}) \leq W(P^*, \mathbf{z}) - \epsilon,
\end{equation*}
and continue solving. In practice, we only need to add a relatively small number of constraints to the optimization model before the solver finds the optimal solution. This is because the user's foil routes are not much greater in weight than the optimal shortest routes in the competition instances. Instances where the user's route differs significantly in weight from the shortest route will take substantially longer to solve.

\section{Results}
To validate the tractability of our method, we evaluated it on the problem instances provided by the competition, consisting of five different maps with five foil routes each. All experiments were conducted on a Windows 11 system with a 13th Gen Intel Core i7-1365U processor and 16 GB RAM. The runtime results on these 25 problem instances are given in \cref{tab:results}.

While the time spent per instance varies between the instances, we were able to solve all of them within 10 seconds. Depending on the size of the map, our method spends 0.5 to 2 seconds setting up the problem, and the remaining time is spent in the solver. The slowest instance to solve was Map 1 with Route 4, where our method spent a total of roughly 8 seconds. Most of the time for this instance was spent in the callback function that makes temporary changes to the graph and solves a shortest route problem.

\paragraph{Competition results} In the final competition evaluation on the held-out test instances, our submission ranked fourth in terms of solution quality. Notably, our method was the fastest on every single test instance, with an average runtime of 9.0 seconds compared to 118.8 seconds for the next-fastest submission.

\begin{table}
\centering
\begin{tabular}{@{}cccccr@{}}
\toprule
\multicolumn{2}{c}{Instance} & \multicolumn{3}{c}{Time spent (s)} & \\
\cmidrule(r){1-2} \cmidrule(r){3-5}
map & route & total & solving & in callback & \# callbacks \\
\midrule
0 & 1 & 3.94 & 1.78 & 0.37 & 134 \\
  & 2 & 4.09 & 1.96 & 0.51 & 167 \\
  & 3 & 6.52 & 4.39 & 2.37 & 330 \\
  & 4 & 5.07 & 2.89 & 1.19 & 192 \\
  & 5 & 4.97 & 2.74 & 0.86 & 179 \\
1 & 1 & 1.23 & 0.78 & 0.22 & 177 \\
  & 2 & 1.01 & 0.44 & 0.12 & 127 \\
  & 3 & 0.96 & 0.44 & 0.13 & 138 \\
  & 4 & 7.88 & 7.35 & 4.59 & 8,879 \\
  & 5 & 0.92 & 0.39 & 0.07 & 137 \\
2 & 1 & 1.42 & 0.62 & 0.10 & 128 \\
  & 2 & 1.55 & 0.87 & 0.44 & 151 \\
  & 3 & 1.28 & 0.58 & 0.16 & 142 \\
  & 4 & 1.57 & 0.88 & 0.27 & 154 \\
  & 5 & 3.44 & 2.65 & 1.09 & 5,688 \\
3 & 1 & 1.06 & 0.50 & 0.08 & 122 \\
  & 2 & 1.08 & 0.52 & 0.09 & 139 \\
  & 3 & 0.93 & 0.36 & 0.05 & 121 \\
  & 4 & 1.15 & 0.61 & 0.27 & 180 \\
  & 5 & 1.59 & 1.05 & 0.41 & 1,585 \\
4 & 1 & 1.37 & 1.00 & 0.54 & 483 \\
  & 2 & 2.73 & 2.27 & 0.96 & 10,832 \\
  & 3 & 0.98 & 0.45 & 0.16 & 162 \\
  & 4 & 0.78 & 0.31 & 0.06 & 126 \\
  & 5 & 0.89 & 0.41 & 0.08 & 127 \\
\bottomrule
\end{tabular}
\caption{Breakdown of time spent on each instance from the competition's train set and the number of callbacks performed (equivalently, the number of constraints added) before proving optimality. Our method solves each instance within 10 seconds.}
\label{tab:results}
\end{table}

\section{Future Work} \label{sec:future-work}

Our method solves a restriction of the real problem to optimality: finding a minimal counterfactual explanation for how the user's route becomes the shortest route. The competition's real problem allowed for some slack between the shortest path and the user's route, however. Therefore, we recommend that our method be extended with a search over target routes in the future. This means that one should enumerate paths $r^t_\text{target}$ close to the user's foil path (they should satisfy $d_r\left(r^t, r^t_\text{target})\right) \leq \delta$), and use our method to find a minimal counterfactual for them. After enumerating, the smallest counterfactual encountered can be returned. We believe this to be a tractable approach because the solver can be warm-started after the initial solve, and previously added constraints can be memorized, which will improve solving time.

\bibliographystyle{named}
\bibliography{sources}

\end{document}